\documentclass{article}
\usepackage{ijcai22}

\usepackage{times}

\usepackage{soul}
\usepackage{url}
\usepackage[hidelinks]{hyperref}
\usepackage[utf8]{inputenc}
\usepackage[small]{caption}
\usepackage{graphicx}
\usepackage{amsmath}
\usepackage{booktabs}
\usepackage{multirow}
\usepackage{graphicx}
\usepackage[ruled]{algorithm2e}
\usepackage{xcolor}

\newcommand*\samethanks[1][\value{footnote}]{\footnotemark[#1]}

\title{Bridging the Gap between Reality and Ideality of Entity Matching:\\  A Revisiting and Benchmark Re-Construction}

\author{
Tianshu Wang$^{1,4}$\and
Hongyu Lin$^1$\thanks{Corresponding author.}\and
Cheng Fu$^1$\and
Xianpei Han$^{1,2,5}$\samethanks\and
Le Sun$^{1,2}$\and \\
Feiyu Xiong$^3$\and
Hui Chen$^3$\and
Minlong Lu$^3$\and
Xiuwen Zhu$^3$
\affiliations
$^1$Chinese Information Processing Laboratory $^2$State Key Laboratory of Computer Science \\
Institute of Software, Chinese Academy of Sciences \\
$^3$Alibaba Group, China \\
$^4$Hangzhou Institute for Advanced Study, University of Chinese Academy of Sciences\\
$^5$Beijing Academy of Artificial Intelligence \\
\emails
\{tianshu2020, hongyu, fucheng, xianpei, sunle\}@iscas.ac.cn, \\
\{feiyu.xfy, weidu.ch, luminlong.lml, xiuwen.zxw\}@alibaba-inc.com
}

\begin{document}
 
\maketitle

\begin{abstract}
Entity matching (EM) is the most critical step for entity resolution (ER). While current deep learning-based methods achieve very impressive performance on standard EM benchmarks, their real-world application performance is much frustrating. In this paper, we highlight that such the gap between reality and ideality stems from the unreasonable benchmark construction process, which is inconsistent with the nature of entity matching and therefore leads to biased evaluations of current EM approaches. To this end, we build a new EM corpus and re-construct EM benchmarks to challenge critical assumptions implicit in the previous benchmark construction process by step-wisely changing the restricted entities, balanced labels, and single-modal records in previous benchmarks into open entities, imbalanced labels, and multi-modal records in an open environment. Experimental results demonstrate that the assumptions made in the previous benchmark construction process are not coincidental with the open environment, which conceal the main challenges of the task and therefore significantly overestimate the current progress of entity matching. The constructed benchmarks and code are publicly released \footnote{{\nolinkurl{https://github.com/tshu-w/ember}}}.
\end{abstract}

\section{Introduction}

\begin{figure}[t]
  \setlength{\belowcaptionskip}{-10pt}
  \centering \includegraphics[width=0.45\textwidth]{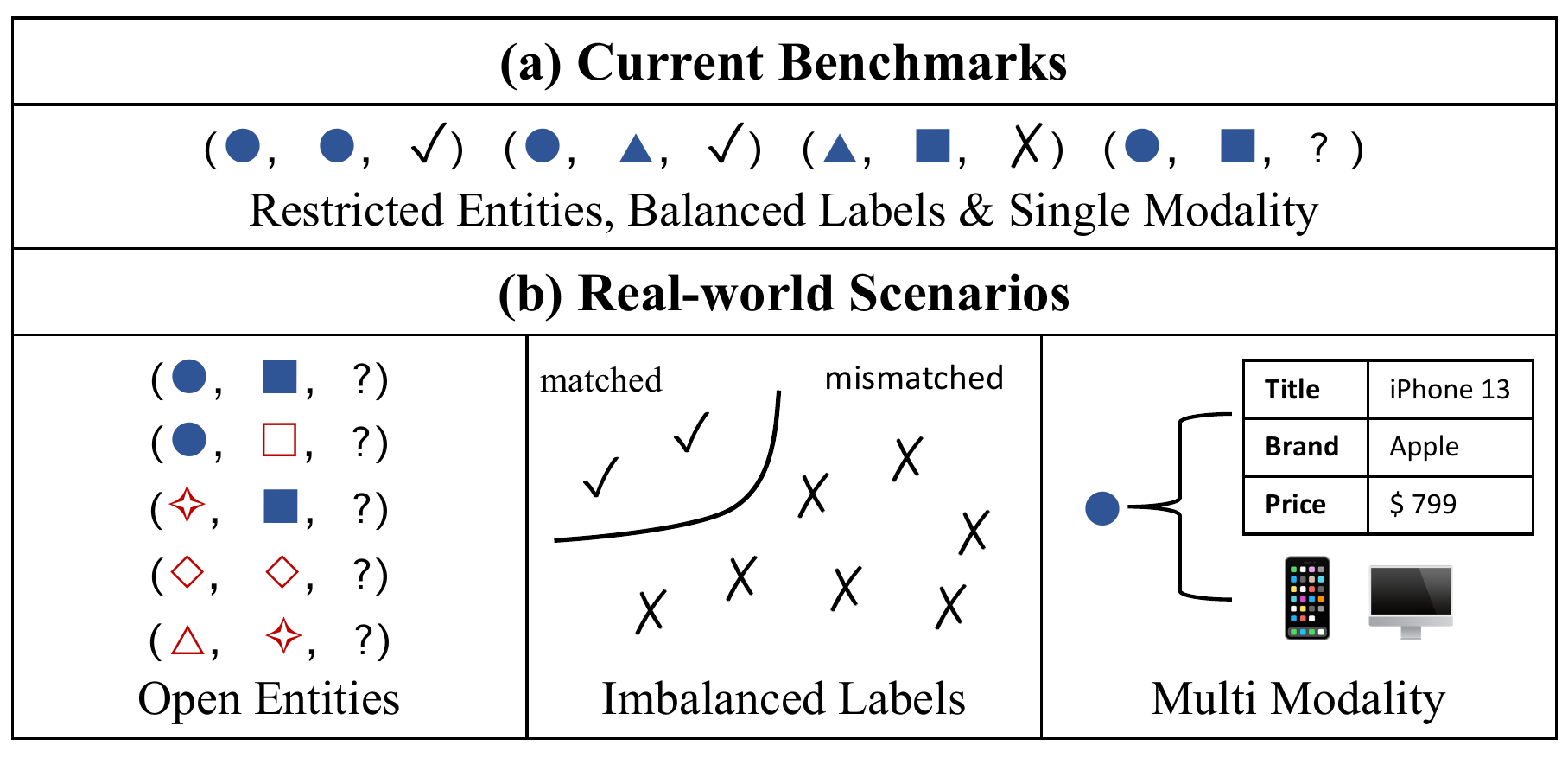}
  \caption{Current benchmarks for EM commonly consider restricted entities, balanced labels, and single modality. However, these conditions do not hold in an open environment, therefore leading to the inconsistency between benchmark performance and real-world applications.}
  \label{fig:introduction}
\end{figure}

Entity resolution, also known as record linkage~\cite{fellegi-69-theor-recor-linkag} or deduplication~\cite{meyer-12-study}, is about merging records that refer to the same real-world entity. Entity matching (EM) aims to identify whether two entity records refer to the same real-world entity, which is the most critical step for entity resolution. For example, an effective entity matcher should resolve the two \emph{entity records} ``iPhone 13 Pro, CA'' and ``apple, iphone, 13 pro'' into the same \emph{entity cluster} because they refer to the same real-world product. Recent years have witnessed the impressive development of EM approaches, especially with the rapid progress of deep learning-based methods~\cite{mudgal-18-deep-learn-entit-match,fu-20-hierar-match-networ-heter-entit-resol,li-20-deep-entit-match-pre-train-languag-model}.
Commonly, these methods are evaluated on the various EM benchmarks~\cite{primpeli-20-profil-entit-match-bench-tasks} from different domains but constructed with a similar process. The state-of-the-art (SOTA) approaches have achieved over 90\% $F_1$ score on most of these datasets, which shows their strong effectiveness on these benchmarks.

However, once launched in real applications, the SOTA EM systems can become frustrating due to the discrepancies between current benchmarks and the open environment. Such discrepancies result in biased scores on the established benchmarks which can not well represent the real-world application performance. Unfortunately, there is no literature looking deep into this reality-ideality gap of entity matching, and therefore causes and effects of these discrepancies are not well studied. As a result, we are unable to evaluate the actual progress of novel approaches to real-world entity matching.

In this paper, we highlight that the discrepancies between benchmarks and the open environment stem from the implicit, unaware assumptions introduced during common benchmark construction. Such erroneous assumptions conceal the main challenges of entity matching, which lead to high performance on current benchmarks. Unfortunately, these assumptions do not hold in real-world scenarios, and therefore discrepancies between benchmarks and the open environment occur. Specifically, as illustrated in Figure~\ref{fig:introduction}, there are three critical erroneous assumptions made when constructing training and test sets of current EM benchmarks:
\begin{itemize}
    \item \textbf{Restricted Entity Assumption,} which means that the entity clusters and/or entity records in the test sets of current benchmarks are mostly covered by records in the training sets, as shown in Table~\ref{tab:overview}. This assumption is made because of the cluster/record sampling strategies applied for current benchmarks. Unfortunately, none of the mainstream entity resolution applications can meet, even be close to, such an assumption, because an effective entity matcher should be able to deal with unseen clusters and records in an open environment. As a result, current benchmarks are unable to evaluate entity matchers in an open environment.
    
    \item \textbf{Balanced Label Assumption,} which means that the ratios of mismatched to matched instances are relatively low and quite close in training and test sets of benchmarks. However, entity matching is an extremely imbalanced problem in real-world applications, and such the mismatched/matched ratio is commonly unknown and diverges significantly. Even with the prior blocking step, the ratio of mismatched record pairs to matched pairs can be up to 100:1~\cite{thirumuruganathan-21-deep}. This poses the critical but ignored challenge to train and evaluate entity matchers in highly-imbalanced settings. As a result, current benchmarks are unable to evaluate entity matchers in the real-world imbalanced situation.
    
    \item \textbf{Single Modality Assumption,} which means current benchmarks mainly focus on textual attributes of entity records. This is due to the absence of the high-quality multi-modal entity corpus. However, in an open environment with noisy textual attributes, information from other modalities like images or audio can play a decisive role in entity matching. As a result, current benchmarks are difficult to be applied to accurately evaluate the effectiveness of multi-modal information for EM.
\end{itemize}

\begin{table}[t]
  \setlength{\belowcaptionskip}{-10pt}
  \centering
  \resizebox{0.48\textwidth}{!}{%
    \begin{tabular}{c|rrr}
    \hline
    Benchmarks               & Matched:    & Seen         & Seen    \\
                            & Mismatched      & Clusters     & Records \\ \hline
    abt-buy~\cite{mudgal-18-deep-learn-entit-match}        & $\approx$1:6           & 99\%         & 96\%    \\
    amazon-google~\cite{mudgal-18-deep-learn-entit-match}  & $\approx$1:6           & 99\%         & 97\%    \\
    dblp-acm~\cite{mudgal-18-deep-learn-entit-match}       & $\approx$1:20          & 100\%        & 100\%   \\
    dblp-scholar~\cite{mudgal-18-deep-learn-entit-match}   & $\approx$1:15          & 100\%        & 100\%   \\
    walmart-amazon~\cite{mudgal-18-deep-learn-entit-match} & $\approx$1:12          & 100\%        & 99\%    \\
    cora~\cite{wang-21-entity}           & $\approx$1:4           & 100\%        & 100\%   \\    wdc\_cameras~\cite{primpeli-19-wdc-train-datas-gold-stand}   & $\approx$1:3           & 100\%        & 78\%    \\
    wdc\_watchers~\cite{primpeli-19-wdc-train-datas-gold-stand}  & $\approx$1:3           & 100\%        & 81\%    \\
    wdc\_computers~\cite{primpeli-19-wdc-train-datas-gold-stand} & $\approx$1:3           & 100\%        & 72\%    \\
    wdc\_shoes~\cite{primpeli-19-wdc-train-datas-gold-stand}     & $\approx$1:3           & 100\%        & 62\%    \\ \hline
    \end{tabular}%
  }
  \caption{Overview of previous EM benchmarks. We omit the datasets in some benchmarks whose test set is less than 100 instances. We can see that they are with very high seen cluster and record ratios, as well as a relatively low mismatched/matched ratio. However, these conditions no longer hold in an open environment.}
  \label{tab:overview}
\end{table}
To assess how these three implicit assumptions bias evaluations of entity matchers, this paper constructs a new multi-modal corpus, which comes from a large Chinese e-commerce website and contains more than 120,000 records for 10,000 products. Each record in the corpus is with a high-quality image attribute. Then starting from the corpus and standard benchmark construction criteria, we re-construct new benchmarks by removing the above-mentioned three erroneous assumptions step-by-step to revisit how these assumptions influence the evaluations on previous benchmarks. Specifically, for restricted entity assumption, we propose to leverage more practical cluster/record sampling strategies to build three benchmarks based on three most common EM application paradigms. For balanced label assumption, we vary the ratio of mismatched to matched instances on training and test set respectively to see how the real-world imbalanced situation can influence the evaluation. For single modality assumption, we thoroughly evaluate the effectiveness of introducing visual attributes based on the above real-world settings. From the newly re-constructed benchmarks, we find previous benchmarks are far from evaluating entity matching in an open environment because
\begin{itemize}
    \item \textbf{Restricted Entity Assumption biases the nature of the task of entity matching.} The assumption changes the task from learning an effective matcher to learning effective representations of \textbf{\textit{seen}} clusters/records. This discrepancy makes previous benchmarks significantly overestimate the performance of current entity matchers.
    
    \item \textbf{Balanced Label Assumption conceals the most critical challenge of entity matching.} We find that in real-world imbalanced scenarios, the performance on imbalanced test sets will dramatically diverge from the evaluation results on previous benchmarks, no matter how the degree of balance of the training set changes.  
    
    \item \textbf{Single Modality Assumption stems from the underestimation of the importance of multi-modality on previous benchmarks.} We find that in the open environment, visual information can improve the performance of entity matching  significantly. Moreover, the importance of visual information can significantly rise especially in open and imbalanced settings. This also confirms that previous restricted entities and balanced label benchmarks can not well estimate the effectiveness of multi-modal entity matchers.
\end{itemize}

Our re-constructed benchmark clearly shows that previous benchmark construction criteria can not cover the main challenges of entity matching, and there is still a long way to build an effective entity matcher in an open environment. Generally speaking, the main contributions of this paper are:

\begin{itemize}
  \item \textbf{We reveal three implicit assumptions behind current EM benchmarks.} These assumptions are introduced during the benchmark construction process, which leads to significant discrepancies between current benchmarks and open environment. 
  \item \textbf{We build a new multi-modal entity matching corpus.} The corpus contains more than 120,000 multi-modal records for 10,000 products. This corpus provides a solid foundation for evaluating the impact of above assumptions, as well as future research on multi-modal EM.
  \item \textbf{We thoroughly assess the impact of the above three critical assumptions for evaluating entity matcher.} Experimental results show that these assumptions conceal the most critical challenges of entity matching, and therefore significantly overestimate the current progress due to the biased benchmarks.
\end{itemize}

\section{Background and New Corpus Construction}

In this section, we will first demonstrate the construction process of conventional entity matching benchmarks and point out three implicit assumptions made during the process. Then to assess the impacts of the three assumptions, we present a new multi-modal corpus for entity matching. Finally, we will briefly introduce the state-of-the-art approaches on entity matching, which will be used to evaluate the impact of the above-mentioned assumptions in the following sections.

\subsection{Construction and Implicit Assumptions of Previous Benchmarks}
Algorithm~\ref{alg:process} outlines the construction process of current entity matching benchmarks.
The procedure commonly first removes non-textual attributes from all records. Then a fixed number of entity clusters are sampled to construct datasets. Some record pairs within the same clusters are regarded as the matched entity record pairs, and a number of mismatched pairs are sampled from records from different clusters. The number of mismatched instances is commonly proportional to the number of matched instances with a fixed ratio. Finally, all record pairs with matched/mismatched labels are split into training/validation/test sets to build the standard benchmarks.

Such a construction process, however, is implicitly incorporated with three assumptions. First, because all records are sampled from the same group of clusters $C'$, there is a great chance that clusters and records in the test set would appear in the training set. To show this, Table~\ref{tab:overview} presents the statistics from several most widely-used EM benchmarks. We can see that nearly all entity clusters in the test set of these benchmarks are covered by the training set. Furthermore, a vast majority of entity records in the test set are also covered by the training set. However, in an open environment this assumption does not hold, because a great number of entity clusters and records are unseen during training. Second, due to the mismatched-matched instance sampling strategy, the ratio of mismatched pairs to matched pairs is relatively low, as shown in Table~\ref{tab:overview} again. Further, such ratios are nearly the same for the training and test set. Unfortunately, in real-world applications we will face an extremely high mismatched-matched ratio. Even after the blocking of entity resolution, a record may have up to 100 candidate matches. And due to the long-tail phenomenon, it is very frequently that only one instance among them is the matched one. Consequently, current benchmarks can not well reflect how well entity matchers can deal with such an extremely imbalanced situation. Third, many current benchmarks only focus on textual attributes, which multi-modal attributes become increasingly popular and important in many EM scenarios. Due to the above reasons, current benchmarks with these implicit assumptions can not well evaluate the performance of the entity matcher in the open environment.

\begin{algorithm}[!t]
  \caption{The Common Construction Process of \textbf{Previous} Entity Matching Benchmarks}\label{alg:process}
  \KwData{A set of entity clusters $C = \{c_1, c_2, \cdots, c_n \}$, where each cluster $c_i$ contains several records $c_i = \{r_i^1, r_i^2, \cdots, r_i^m\}$}
  % \KwIn{train/val/test ratio $r$, mismatched/matched ratio $k$}
  \KwIn{train/val/test ratio, mismatched/matched ratio: $r, k$}
  \KwOut{training, validation, and test set: $train, val, test$}
  \ForEach{$c_i \in C$}{
    \ForEach{$r_i^j \in c_i$}{
      \textbf{\textcolor{red}{\textasteriskcentered\textasteriskcentered Single modality assumption\textasteriskcentered\textasteriskcentered}} \\
      preserving only textual attributes of record $r_i^j$ \\
    }
  }
  sample a subset of clusters $C^\prime$ from $C$ \\
  $D = \O$ \\
  \ForEach{$c_i \in C^\prime$}{
    \For{$r_i^j, r_i^k \in c_i$}{
      $D \gets D \cup (r_i^j, r_i^k, \text{matched})$ \\
    }
  }
  \textbf{\textcolor{red}{\textasteriskcentered\textasteriskcentered Balanced label assumption\textasteriskcentered\textasteriskcentered}} \\
  \For{$i \gets 1$ \KwTo $k * \|D\|$}{
    % sample two different clusters $c_i$ and $c_j$ from $C\prime$ \\
	% sample a record $r_1$ from $c_i$ \\
	% sample a record $r_2$ from $c_j$ \\
    sample different cluster records $r_i^l$, $r_j^m$ from $C^\prime$ \\
	$D \gets D \cup (r_i^l, r_j^m, \text{mismatched})$ \\
  }
  \textbf{\textcolor{red}{\textasteriskcentered\textasteriskcentered Restricted entity assumption\textasteriskcentered\textasteriskcentered}} \\ 
  $train, val, test \gets \textsc{random\_split}(D, r)$ \\
  \Return $train, val, test$
\end{algorithm}

\subsection{New Corpus Construction}

In order to assess how these three implicit assumptions influence the evaluations of entity matchers, we construct a new multi-modal corpus based on product information from a Chinese e-commerce website. The newly built corpus contains 3 main categories (clothing, shoes, and accessories) of products. Each product record includes a record ID, title, categorical info, cluster ID, attribute pairs (e.g., color, style, material, etc.), as well as a product image. Entity records with the same cluster ID, which are manually checked, are considered as referring to the same real-world product entity. Finally, there are 10,000 products and 126,277 records in this corpus. Each product has records between 10 and 20. 

From the corpus, we use the clusters from all categories and three specific categories to build the datasets. For each dataset, we randomly sample 250 clusters to build the training set and hold out 100 clusters for the unseen cluster benchmark. For each cluster in the training set, we also hold out 40\% of the records as unseen records.
%\red{The detailed statistics of the datasets are shown in the Appendix.}

\subsection{Baselines}
In this paper, we mainly focus on exploring the impact of the three above problems on deep learning-based approaches while leaving others for future work. We choose two representative methods as baselines in our experiments:
%In order to explore the impact of the three above problems on EM, we choose the two representative entity matching approaches as baselines in our experiments:

\begin{itemize}
  \item \textbf{Deepmatcher (DMatcher)}, which is the first detailed exploration of deep learning methods on EM~\cite{mudgal-18-deep-learn-entit-match}. We use their open source code directly.
  \item \textbf{Ditto}, which is the SOTA EM method based on pre-trained language models~\cite{li-20-deep-entit-match-pre-train-languag-model}. We reproduce and obtain comparable results on existing benchmarks.
\end{itemize}

\section{Benchmark Reconstruction for EM}

\begin{figure}[t]
  \setlength{\belowcaptionskip}{-10pt}
  \centering \includegraphics[width=0.45\textwidth]{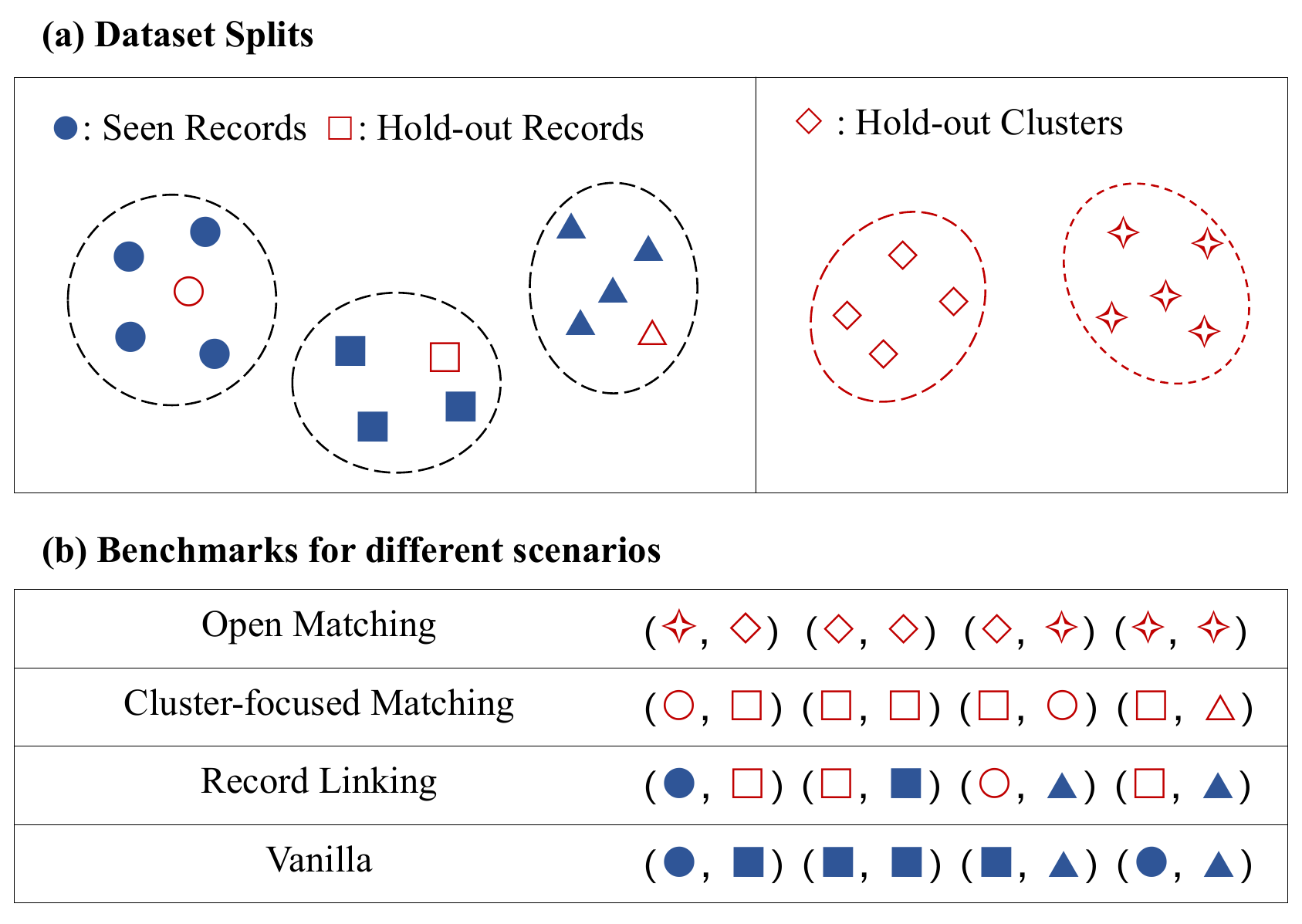}
  \caption{Four kinds of benchmarks we constructed: a) Open Matching, which contains records all from unseen clusters; b) Cluster-focused Matching, which contains unseen records all from seen clusters; c) Record Linking, which contains one seen record and one unseen record from seen clusters; d) Vanilla setup, which contains records that are all seen during training.}
  \label{fig:benchmarks}
\end{figure}

Applications of entity matching in the open environment are highly diversified and complicated. So it is difficult to build one benchmark for all downstream EM applications. To this end, this paper investigates the three most representative applications of entity matching and evaluates how well the current SOTA approaches can accomplish these applications. Specifically, we build benchmarks for the following three kinds of typical downstream paradigms of entity matching:
\begin{itemize}
    \item \textbf{Open Matching (OM),} which tries to identify whether two open records refer to the same entity clusters, and there is no specific restriction for the entity clusters. This paradigm corresponds to the scenarios where we want to leverage the learned entity matcher as a universal tool for merging or deduplicating (e.g., combining two databases, deduplicating a new table, etc.). As a result, the clusters and records in the test benchmark are commonly unknown during training under this scenario.
    \item \textbf{Cluster-focused Matching (CFM),} which tries to identify whether two records belonging to specific \emph{seen} clusters are matched. This paradigm corresponds to the scenarios where we try to sort the mined records about a group of products or merge information about specific entities from multiple sources. Under this scenario, all entity clusters in the test set are observed during training, but all records in the test benchmark are new and unseen.
    \item \textbf{Record Linking (RL),} which tries to link a newly-obtained record to other records in the database. Under this scenario, we can use all entity clusters in the database for training, and there is no new cluster in the test set. However, for each record pair to classify, only one record in it has appeared in the training set, and another record is new. This is a very common scenario for entity matching applications, e.g., linking a new product record to the constructed commodity library.
\end{itemize}

Apart from the above three benchmarks, we also build a vanilla benchmark for comparison, which follows the construction criteria of the WDC benchmarks~\cite{primpeli-19-wdc-train-datas-gold-stand}. Figure~\ref{fig:benchmarks} illustrates how we build these benchmarks upon our constructed corpus. All these 4 benchmarks share the same training and validation sets. For the test benchmark of OM, we apply the same record pair generation procedure as Algorithm~\ref{alg:process} on the \emph{hold-out clusters} to generate matched and mismatched record pairs. For CFM, we use the \emph{hold-out records} from the training set to build the test benchmark. For RL, we first sample a record from the \emph{hold-out records} and then sample a matched or mismatched record to it from the training set. We set the default matched-mismatched ratio on training and test sets to 1:3 like the WDC\footnote{\href{http://webdatacommons.org/largescaleproductcorpus/v2/\#toc4.2}{\nolinkurl{webdatacommons.org/largescaleproductcorpus/v2/\#toc4.2}}}. However, in experiments, we also vary the matched-mismatched ratio of the benchmarks to investigate how the balanced label assumption influences EM performance. Besides, following the same criteria, we also build these 4 benchmarks for each category. 

\section{Experiments and Findings}
\subsection{Restricted Entity Assumption}

\begin{table}[!t]
  \setlength{\belowcaptionskip}{-10pt}
  \centering
  \resizebox{0.4\textwidth}{!}{%
    \begin{tabular}{@{}cc|c|ccc@{}}
    \toprule
                                 &       & Vanilla & RL & CFM & OM \\ \midrule
    \multirow{2}{*}{All}         & DMatcher    & 78.48       & 70.20      & 65.18       & 53.09              \\
                                 & Ditto & 93.06       & 88.27      & 84.02       & 67.82              \\ \midrule
    \multirow{2}{*}{Cloth.}    & DMatcher    & 77.62       & 69.02      & 63.43       & 57.40              \\
                                 & Ditto & 89.75       & 82.84      & 80.20       & 70.54              \\ \midrule
    \multirow{2}{*}{Shoes}       & DMatcher    & 75.44       & 69.09      & 63.60       & 54.49              \\
                                 & Ditto & 84.93       & 82.08      & 76.57       & 62.21              \\ \midrule
    \multirow{2}{*}{Acc.} & DMatcher    & 81.47       & 70.73      & 63.16       & 50.67              \\
                                 & Ditto & 91.41       & 84.93      & 81.23       & 61.19              \\ \bottomrule
    \end{tabular}%
  }
  \caption{$F_1$ scores on 4 benchmarks. We can see that unseen clusters and records can significantly reduce the performance of entity matchers.}
  \label{tab:assp_1}
\end{table}

\textbf{Findings 1. } \emph{Restricted entity assumption biases the nature of entity matching, which changes the task from learning an effective matcher to learning effective representations of seen clusters/records.}

To demonstrate this, we compared the performance of the SOTA entity matchers on 4 newly constructed benchmarks. Table~\ref{tab:assp_1} shows the results. We can find that on the vanilla benchmark, these methods can achieve competitive performance of nearly 90\% $F_1$ scores, which is similar to that on previous benchmarks. Unfortunately, the performance dramatically dropped when the restricted entity assumption was removed. We find that the model performance dropped most significantly in the realistic Open Matching scenario -- nearly 30\% of $F_1$ scores. Meanwhile, the performance also have a significant drop in Record Linking and Cluster-focused Matching, even most clusters or records in these two settings have been observed during training. This verifies that previous benchmarks incorporated with restricted entity assumption significantly overestimate the performance. Furthermore, we find that the more information about seen records or clusters in benchmarks, the better the performance of the models. This shows that current models, trained on standard benchmarks, do not learn sufficient information to build a universal matcher. Instead, they pay more attention to learning sufficient representations of seen clusters and records. However, in order for the entity matcher can be used in a broader range of scenarios, we hope that the learned entity matcher can be generalized to unseen clusters and records. As a result, only focusing on learning representations of seen clusters or records is not sufficient for EM in an open environment.

In general, previous benchmarks mainly evaluate the ability to learn on seen records or clusters but cannot reflect the ability of generalized entity matching in the open environment. Therefore, there is a discrepancy between previous benchmarks and the real-world entity matching applications.

\subsection{Balanced Label Assumption}

\begin{table}[!t]
  \setlength{\belowcaptionskip}{-10pt}
  \centering
  \resizebox{0.4\textwidth}{!}{%
    \begin{tabular}{@{}cc|c|ccc@{}}
    \toprule
                            &          & Vanilla & RL    & CFM   & OM    \\ \midrule
    \multirow{2}{*}{All}    & DMatcher & 16.43   & 13.33 & 11.93 & 6.98  \\
                            & Ditto    & 38.52   & 32.71 & 28.55 & 14.52 \\ \midrule
    \multirow{2}{*}{Cloth.} & DMatcher & 15.70   & 11.65 & 10.11 & 8.02  \\
                            & Ditto    & 33.39   & 27.15 & 24.86 & 18.15 \\ \midrule
    \multirow{2}{*}{Shoes}  & DMatcher & 13.91   & 10.77 & 9.59  & 6.31  \\
                            & Ditto    & 24.53   & 21.90 & 20.39 & 10.11 \\ \midrule
    \multirow{2}{*}{Acc.}   & DMatcher & 16.85   & 13.89 & 11.86 & 6.84  \\
                            & Ditto    & 30.44   & 25.24 & 23.18 & 10.58 \\ \bottomrule
    \end{tabular}%
  }
  \caption{$F_1$ scores on 4 benchmarks with mismatched/matched ratio = 100. We find that the performance dramatically dropped on imbalanced benchmarks compared with balanced benchmarks in Table~\ref{tab:assp_1}.}
  \label{tab:assp_2}
\end{table}

\textbf{Findings 2. } \emph{Balanced label assumption conceals the most critical challenge of entity matching.}

\begin{figure}[!t]
  \setlength{\belowcaptionskip}{-10pt}
  \centering
  \includegraphics[width=0.4\textwidth]{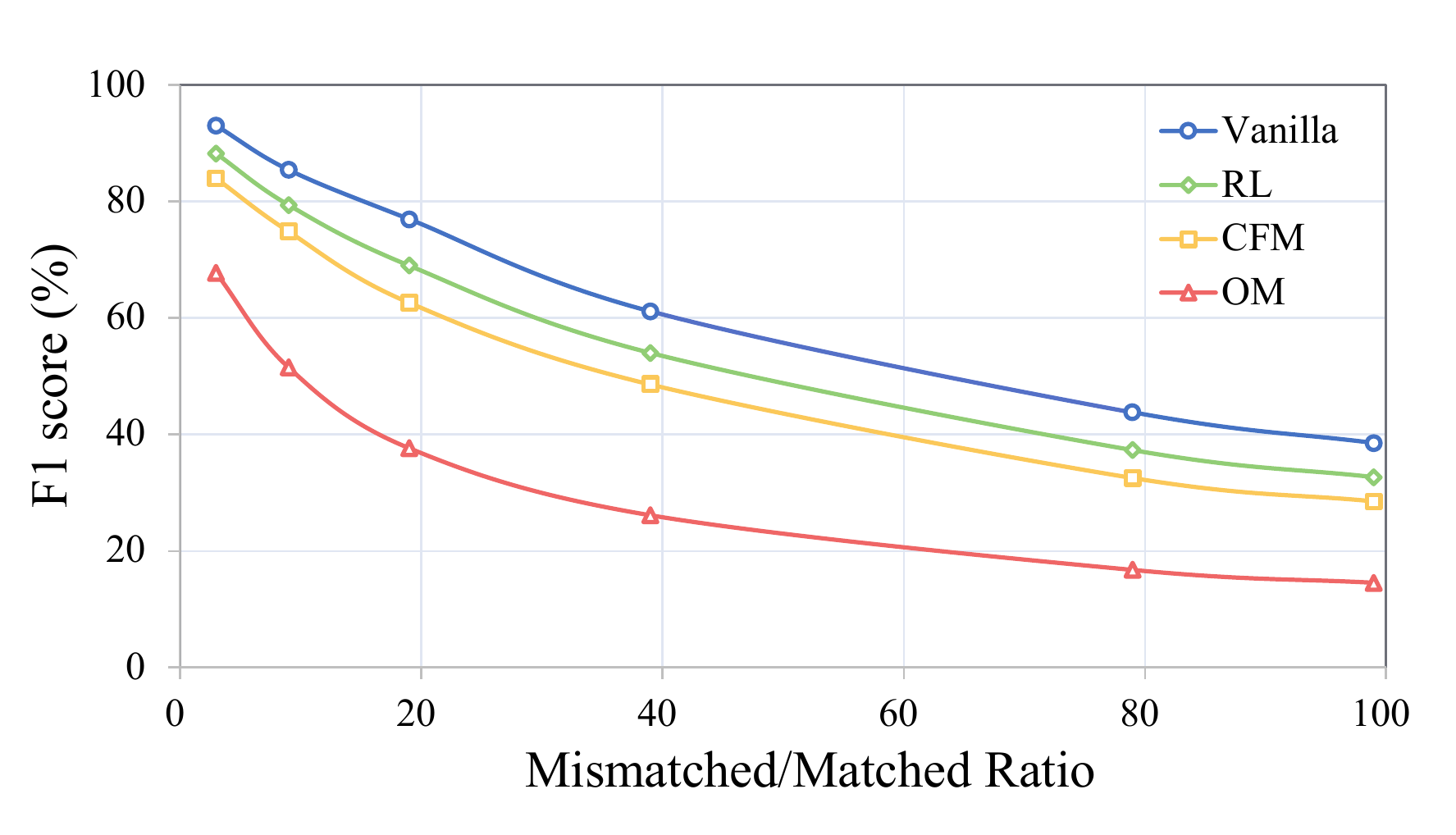}
  \caption{$F_1$ scores on 4 benchmarks w.r.t. different ratios of mismatched-matched ratio on \emph{test set}.}
  \label{fig:imbalance_test}
\end{figure}

\begin{figure}[!t]
  \setlength{\belowcaptionskip}{-10pt}
  \centering
  \includegraphics[width=0.4\textwidth]{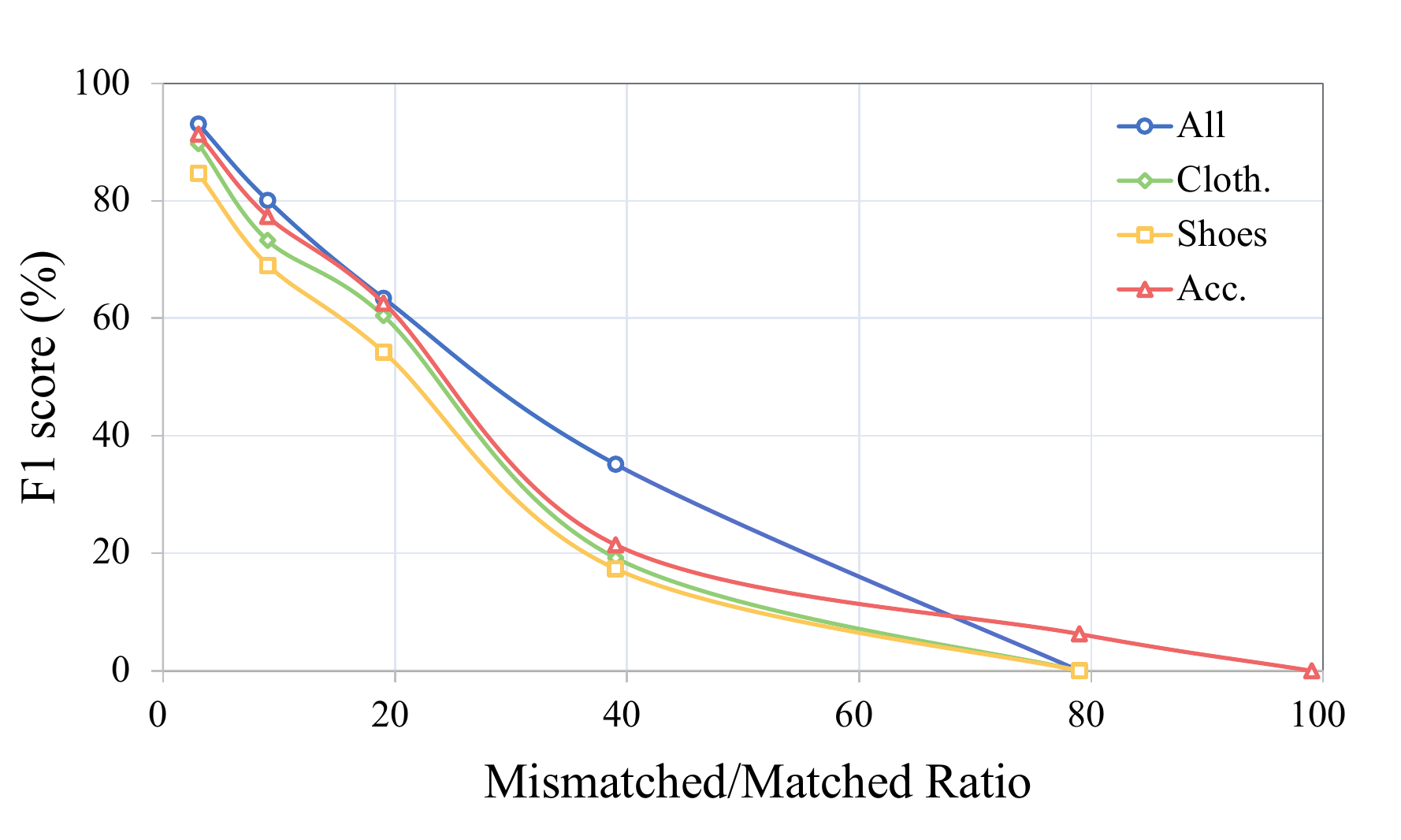}
  \caption{$F_1$ scores on 4 categories w.r.t. different ratios of mismatched-matched ratio on \emph{both training and test sets}.}
  \label{fig:imbalance_train_test}
\end{figure}

To show this, we investigated the performance of the baselines on 4 newly constructed benchmarks and varied the ratio of matched record pairs to mismatched record pairs gradually from 1:3 to 1:100. Note that 1:100 is a reasonable matched-mismatched ratio in real-world entity matching applications, because the entity resolution system may keep more than 100 candidates for a record at the blocking step to ensure recall~\cite{thirumuruganathan-21-deep}. And due to the long-tail phenomenon, it is very frequently that only one instance among them is the matched pair. So a matched-mismatched ratio of 1:100 corresponds to real-world EM applications.

Table~\ref{tab:assp_2} shows the results. We can see that even on the vanilla benchmark, the $F_1$ score dropped dramatically on the imbalanced scenario, compared to the results on the balanced scenario in Table~\ref{tab:assp_1}. More importantly, the performance becomes even worse in other three benchmarks. On the open matching benchmark, the performance of state-of-the-art systems is even as low as 14.52\%. This shows the significant impact of imbalanced labels on entity matching.

In order to take a closer look at the challenges posed by imbalanced labels, we dynamically varied the mismatched-matched ratio on test benchmark. Figure~\ref{fig:imbalance_test} shows the results. We can see that the performance on all benchmarks steadily dropped when the mismatched-matched ratio increased on the test benchmark. Furthermore, we find that this problem can not be solved by merely adjusting the mismatched-matched ratio on the training set, as shown in Figure~\ref{fig:imbalance_train_test}. We believe that this is because the imbalanced training set will pose a great challenge for model learning. As a result, how to deal with the extremely imbalanced labels in the open environment is one of the most critical challenges for entity matching. Unfortunately, previous benchmarks conceal this challenge because they introduced the balanced label assumption when generating mismatched instances. Consequently, previous benchmarks can not well represent the real-world performance of entity matchers in an open environment.

\subsection{Single Modality Assumption}

\begin{table}[!t]
  \setlength{\belowcaptionskip}{-10pt}
  \centering
  \resizebox{0.43\textwidth}{!}{%
    \begin{tabular}{@{}cc|c|ccc@{}}
    \toprule
                              &             & Vanilla & RL & CFM & OM \\ \midrule
    \multicolumn{6}{c}{Balanced (Matched:Mismatched = 1:3)}                                                                               \\ \midrule
    \multirow{3}{*}{All}      & Textual        & 93.06       & 88.27      & 84.02       & 67.82              \\
                              & Visual      & 95.42       & 91.50      & 88.13       & 74.14              \\
                              & Multi-modal & 96.89       & 93.66      & 91.11       & 78.45              \\ \midrule
    \multirow{3}{*}{Cloth.} & Textual        & 89.75       & 82.84      & 80.20       & 70.54              \\
                              & Visual      & 95.87       & 92.08      & 88.12       & 78.12              \\
                              & Multi-modal & 96.90       & 93.40      & 90.18       & 80.56              \\ \midrule
    \multirow{3}{*}{Shoes}    & Textual        & 84.93       & 82.08      & 76.57       & 62.21              \\
                              & Visual      & 89.50       & 83.82      & 79.79       & 64.90              \\
                              & Multi-modal & 91.82       & 87.13      & 85.29       & 72.02              \\ \midrule
    \multirow{3}{*}{Acc.}     & Textual        & 91.41       & 84.93      & 81.23       & 61.19              \\
                              & Visual      & 92.64       & 88.48      & 83.59       & 65.12              \\
                              & Multi-modal & 94.96       & 91.26      & 86.98       & 68.29              \\ \midrule\midrule
    \multicolumn{6}{c}{Imbalanced (Matched:Mismatched = 1:100)}                                                                              \\ \midrule
    \multirow{3}{*}{All}      & Textual        & 38.52       & 32.71      & 28.55       & 14.52              \\
                              & Visual      & 65.82       & 55.92      & 50.29       & 30.93              \\
                              & Multi-modal & 76.72       & 64.67      & 60.06       & 30.39              \\ \midrule
    \multirow{3}{*}{Cloth.} & Textual        & 33.39       & 27.15      & 24.86       & 18.15              \\
                              & Visual      & 65.10       & 54.93      & 49.64       & 29.83              \\
                              & Multi-modal & 73.16       & 62.37      & 58.38       & 30.86              \\ \midrule
    \multirow{3}{*}{Shoes}    & Textual        & 24.53       & 21.90      & 20.39       & 10.11              \\
                              & Visual      & 38.52       & 29.25      & 24.15       & 12.09              \\
                              & Multi-modal & 46.95       & 35.88      & 30.07       & 14.61              \\ \midrule
    \multirow{3}{*}{Acc.}     & Textual        & 30.44       & 25.24      & 23.18       & 10.58              \\
                              & Visual      & 46.54       & 38.85      & 33.36       & 14.93              \\
                              & Multi-modal & 49.02       & 41.24      & 36.03       & 14.37              \\ \bottomrule
    \end{tabular}%
  }
  \caption{Experimental results on multi-modal EM. We can see that introducing the visual attribute can significantly boost the performance on open clusters and imbalanced settings.}
  \label{tab:multi-modal}
\end{table}

\textbf{Findings 3. } \emph{Single modality assumption stems from the underestimation of the importance of multi-modality on previous benchmarks.}

To show this, we conducted experiments on newly constructed benchmarks with multi-modal records. Because currently there is little previous work focused on multi-modal entity matching, we build a simple baseline for visual and multi-modal EM. Specifically, we use Vision Transformer~\cite{alexey-21-vit} as the image encoder and apply a multilayer perceptron on the representations of images of two records to obtain a visual matching representation. Then for the single-modal visual baseline, we direct send the visual matching representation into a classifier to determine the match result. For the multi-modal approach, we use a gated mechanism to fuse this visual matching representation with the text matching representation and then send it to a classifier. 

Table~\ref{tab:multi-modal} shows the results. We can find that the importance of visual attributes can be underestimated based on the performance of the vanilla benchmarks. In balanced vanilla benchmarks, the improvement of introducing visual information is not very significant, which is consistent with the phenomenon from previous benchmarks~\cite{wilke-21-towar-multi-entit-resol-produc-match}. However, when we refer to the results on open environment benchmarks, the improvement of incorporating visual information is very significant: there are more than 7 points of  $F_1$ score gains on CFM and more than 11 points of $F_1$ score gains on OM. Furthermore, the multi-modal model achieves more than 40\% of $F_1$ scores improvements under the imbalanced benchmarks for some categories and can be better generalized to unseen clusters and records. All these results demonstrate that multi-modal information can significantly benefit entity matching in the open environment.

\section{Related Work}

\paragraph{EM Approaches.} Entity matching (EM) aims to identify whether two entity records refer to the same real-world entity, which is the most critical step of entity resolution~\cite{christophides-20-overv-end-end-entit-resol-big-data}. This study dates back to ~\cite{fellegi-69-theor-recor-linkag} and has attached great attention. To solve this open problem, various approaches have been proposed, e.g., distance-based, rule-based, declarative, and probabilistic methods~\cite{papadakis-21-four-gener-entit-resol}. In recent years, deep learning has been introduced to this field and achieved promising results~\cite{thirumuruganathan-18-reuse-adapt-entit-resol-trans-learn,mudgal-18-deep-learn-entit-match,nie-19-deep-sequen-sequen-entit-match,fu-20-hierar-match-networ-heter-entit-resol,li-20-deep-entit-match-pre-train-languag-model}. % ge-21-collab,zhao-19-auto-em,shao-20-ergan
\paragraph{EM Benchmarks.} 
In the early development of EM, many datasets are used to construct benchmarks to evaluate EM methods~\cite{primpeli-20-profil-entit-match-bench-tasks}. There has been an effort on building large-scale datasets for deep learning methods~\cite{primpeli-19-wdc-train-datas-gold-stand}. There have also been some attempts on extending the EM task to broader scenarios by extending the data schema, record formats, and relationships between records~\cite{jim-20-semtab,wang-21-macham}. Unfortunately, as we mentioned above, all these benchmarks are built on three erroneous assumptions, which lead to a significant gap between the benchmarks and EM in the open environment. There is also some literature discussing multi-modal entity matching~\cite{christophides-20-overv-end-end-entit-resol-big-data,wilke-21-towar-multi-entit-resol-produc-match}. However, due to the benchmark limitation, the importance and effectiveness of multi-modal attributes to EM were hindered and inaccurately evaluated.

\section{Conclusions}
In this paper, we highlight that the gap between reality and ideality of entity matching stems from the erroneous implicit assumptions introduced during the benchmark construction process. These assumptions are inconsistent with the nature of entity matching and therefore lead to biased evaluations of current EM approaches. To this end, we build a new EM corpus and re-construct EM benchmarks. By step-wisely changing the restricted entities, balanced labels, and single-modal records in previous benchmarks into open entities, imbalanced labels, and multi-modal records in an open environment, we find that current state-of-the-art approaches suffer severely from unseen clusters, imbalanced labels. Furthermore, previous benchmarks also underestimated the impact of multi-modal attributes on entity matching.
Our findings reveal that previous benchmarks biased the evaluation of the progress of current entity matching approaches, and there is still a long way to go to build effective entity matchers.

\section*{Acknowledgments}
This work was supported by the National Key Research and Development Program of China (No. 2020AAA0106400), and the National Natural Science Foundation of China under Grants no. 62122077, 62106251, 62076233.

\newpage

%% The file named.bst is a bibliography style file for BibTeX 0.99c
\bibliographystyle{named}
\bibliography{main}

\end{document}